\documentclass[10pt,twocolumn,letterpaper]{article}

\usepackage[pagenumbers]{arxiv} 

\usepackage{graphicx}
\usepackage{amsmath}
\usepackage{amssymb}
\usepackage{booktabs}
\usepackage{cuted}
\usepackage{makecell}
\usepackage{bbding}
\usepackage{mathbbol}
\usepackage{amsmath}
\usepackage{multirow}

\usepackage{tabularx} 
\usepackage{ragged2e} 
\usepackage{booktabs} 
\newcolumntype{L}{>{\RaggedRight\hangafter=1\hangindent=0em}X}

\usepackage[marginal]{footmisc}

\usepackage[pagebackref,breaklinks,colorlinks]{hyperref}

\usepackage[capitalize]{cleveref}
\crefname{section}{Sec.}{Secs.}
\Crefname{section}{Section}{Sections}
\Crefname{table}{Table}{Tables}
\crefname{table}{Tab.}{Tabs.}
\newcommand{\myparagraph}[1]{\vspace{0.1em}\noindent\textbf{#1}}

\begin{document}

\title{NeuralDome: A Neural Modeling Pipeline on Multi-View Human-Object Interactions}


\author {
    Juze Zhang \textsuperscript{\rm 1,\rm 2,\rm 3,\rm 4,\rm *},
    Haimin Luo \textsuperscript{\rm 1,\rm 4,\rm *},
    Hongdi Yang \textsuperscript{\rm 1,\rm 4},
    Xinru Xu \textsuperscript{\rm 1,\rm 4},
    Qianyang Wu \textsuperscript{\rm 1,\rm 4},
    Ye Shi \textsuperscript{\rm 1,\rm 4}, \\
    Jingyi Yu \textsuperscript{\rm 1,\rm 4},
    Lan Xu \textsuperscript{\rm 1,\rm 4, \dag},
    Jingya Wang \textsuperscript{\rm 1,\rm 4,\dag} \\
    \textsuperscript{\rm 1} ShanghaiTech University  
    \textsuperscript{\rm 2} Shanghai Advanced Research Institute, Chinese Academy of Sciences\\
    \textsuperscript{\rm 3} University of Chinese Academy of Sciences \\
    \textsuperscript{\rm 4}  Shanghai Engineering Research Center of Intelligent Vision and Imaging \\
    \{zhangjz,luohm,yanghd,xuxr2022,wuqy2022,shiye,yujingyi,xulan1,wangjingya\}@shanghaitech.edu.cn
}



\maketitle
\footnote{* These authors contributed equally.}
\footnote{\dag Corresponding author.}

\begin{strip}\centering
\includegraphics[scale = 0.47]{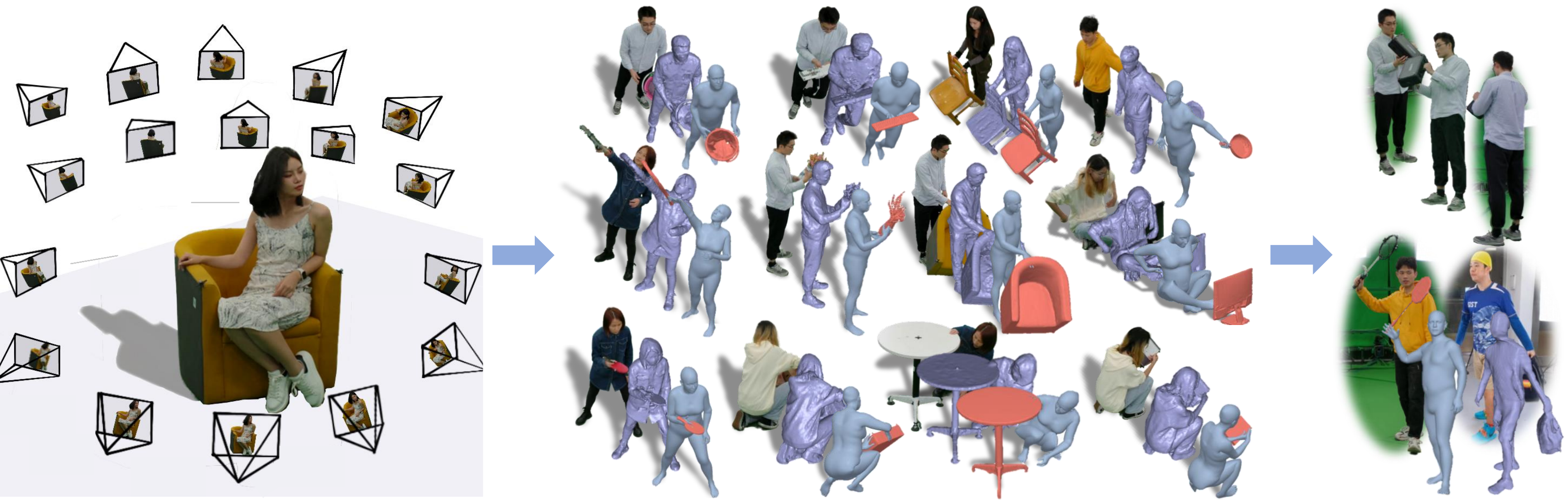}
\captionof{figure}{Our NeuralDome pipeline for processing multi-view video sequences on human object interactions. NeuralDome supports tracking, modeling, and rendering of individual human subjects and objects. To validate NeuralDome, we collect a large dataset HODome over a total of 75 million video frames across 76 viewpoints and process the datasets using NeuralDome for a variety of inference and neural modeling and rendering tasks.}
\label{fig1:teaser}
\end{strip}



\begin{abstract}
Humans constantly interact with objects in daily life tasks. Capturing such processes and subsequently conducting visual inferences from a fixed viewpoint suffers from occlusions, shape and texture ambiguities, motions, etc. To mitigate the problem, it is essential to build a training dataset that captures free-viewpoint interactions. We construct a dense multi-view dome to acquire a complex human object interaction dataset, named HODome, that consists of $\sim$75M frames on 10 subjects interacting with 23 objects. To process the HODome dataset, we develop NeuralDome, a layer-wise neural processing pipeline tailored for multi-view video inputs to conduct accurate tracking, geometry reconstruction and free-view rendering, for both human subjects and objects. Extensive experiments on the HODome dataset demonstrate the effectiveness of NeuralDome on a variety of inference, modeling, and rendering tasks. Both the dataset and the NeuralDome tools will be disseminated to the community for further development. 
\end{abstract}


\section{Introduction} \label{sec:intro}

A key task of computer vision is to understand how humans interact with the surrounding world, by faithfully capturing and subsequently reproducing the process via modeling and rendering. Successful solutions benefit broad applications ranging from sports training to vocational education, digital entertainment to tele-medicine.


Early solutions~\cite{collet2015high,schoenberger2016sfm,motion2fusion} that reconstruct dynamic meshes with per-frame texture maps are time-consuming and vulnerable to occlusions or lack of textures. 
Recent advances in neural rendering~\cite{mildenhall2020nerf,tewari2021advances,Wu_2020_CVPR} bring huge potential for human-centric modeling. Most notably, the variants of Neural Radiance Field (NeRF)~\cite{mildenhall2020nerf} achieve compelling novel view synthesis, which can enable real-time rendering performance~\cite{suo2021neuralhumanfvv,mueller2022instant,wang2022fourier} even for dynamic scenes~\cite{STNeRF_SIGGRAPH2021,peng2021neural,tretschk2021non}, and can be extended to the generative setting without per-scene training~\cite{wang2021ibrnet,kwon2021neural,zhao2022humannerf}. 
However, less attention is paid to the rich and diverse interactions between humans and objects, mainly due to the severe lack of dense-view human-object datasets.
Actually, existing datasets of human-object interactions are mostly based on optical markers~\cite{taheri2020grab} or sparse RGB/RGBD sensors~\cite{liu2019ntu,bhatnagar2022behave}, without sufficient appearance supervision for neural rendering tasks. As a result, the literature on neural human-object rendering~\cite{sun2021HOIFVV,jiang2022neuralhofusion} is surprisingly sparse, let alone further exploring the real-time or generative directions.
Besides, existing neural techniques~\cite{STNeRF_SIGGRAPH2021,shuai2022novel} suffer from tedious training procedures due to the human-object occlusion, and hence infeasible for building a large-scale dataset.
In a nutshell, despite the recent tremendous thriving of neural rendering, the lack of both a rich dataset and an efficient reconstruction scheme constitute barriers in human-object modeling.

In this paper, we present \textit{NeuralDome}, a neural pipeline that takes multi-view dome capture as inputs and conducts accurate 3D modeling and photo-realistic rendering of complex human-object interaction. As shown in Fig.~\ref{fig1:teaser}, NeuralDome exploits layer-wise neural modeling to produce rich and multi-modality outputs including the geometry of dynamic human, object shapes and tracked poses, as well as a free-view rendering of the sequence. 

Specifically, we first capture a novel human-object dome (HODome) dataset that consists of 274 human-object interacting sequences, covering 23 diverse 3D objects and 10 human subjects (5 males and 5 females) in various apparels. We record multi-view video sequences of natural interactions between the human subjects and the objects where each sequence is about 60s in length using a dome with 76 RGB cameras, resulting in ~75 million video frames. We also provide an accurate pre-scanned 3D template for each object and utilize sparse optical markers to track individual objects throughout the sequences. 

To process the HODome dataset, we adopt an extended Neural Radiance Field (NeRF) pipeline. The brute-force adoption of off-the-shelf neural techniques such as Instant-NSR~\cite{zhao2022human} and Instant-NGP~\cite{mueller2022instant}, although effective, do not separate objects from human subjects and therefore lack sufficient fidelity to model their interactions. We instead introduce a layer-wise neural processing pipeline.
Specifically, we first perform a joint optimization based on the dense inputs for accurately tracking human motions using the parametric SMPL-X model~\cite{pavlakos2019expressive} as well as localizing objects using template meshes. We then propose an efficient layer-wise neural rendering scheme where the humans and objects are formulated as a pose-embedded dynamic NeRF and a static NeRF with tracked 6-DoF rigid poses, respectively. Such a layer-wise representation effectively exploits temporal information and robustly tackles the occluded regions under interactions. 
We further introduce an object-aware ray sampling strategy to mitigate artifacts during layer-wise training, as well as template-aware geometry regularizers to enforce contact-aware deformations. Through weak segmentation supervision, we obtain the decoupled and occlusion-free appearances for both the humans and the objects at a high fidelity amenable for training the input multi-view inputs for a variety of tasks from monocular motion capture to free-view rendering from sparse multi-view inputs. 
%

To summarize, our main contributions include:
\begin{itemize} 
	\setlength\itemsep{0em}
	
	\item We introduce NeuralDome, a neural pipeline, to accurately track humans and objects, conduct layer-wise geometry reconstruction, and enable novel-view synthesis, from multi-view HOI video inputs.
	
	\item We collect a comprehensive dataset that we call HODome that will be disseminated to the community, with both raw data and the output modalities including separated geometry and rendering of individual objects and human subjects, their tracking results, free-view rendering results, etc.
	
	\item We demonstrate using the dataset to train networks for a variety of visual inference tasks with complex human object interactions. 
	
	
	
\end{itemize}

\section{Related Works}
\label{sec:formatting}

\begin{figure}[t]
\centering
	\includegraphics[width=1.0\columnwidth]{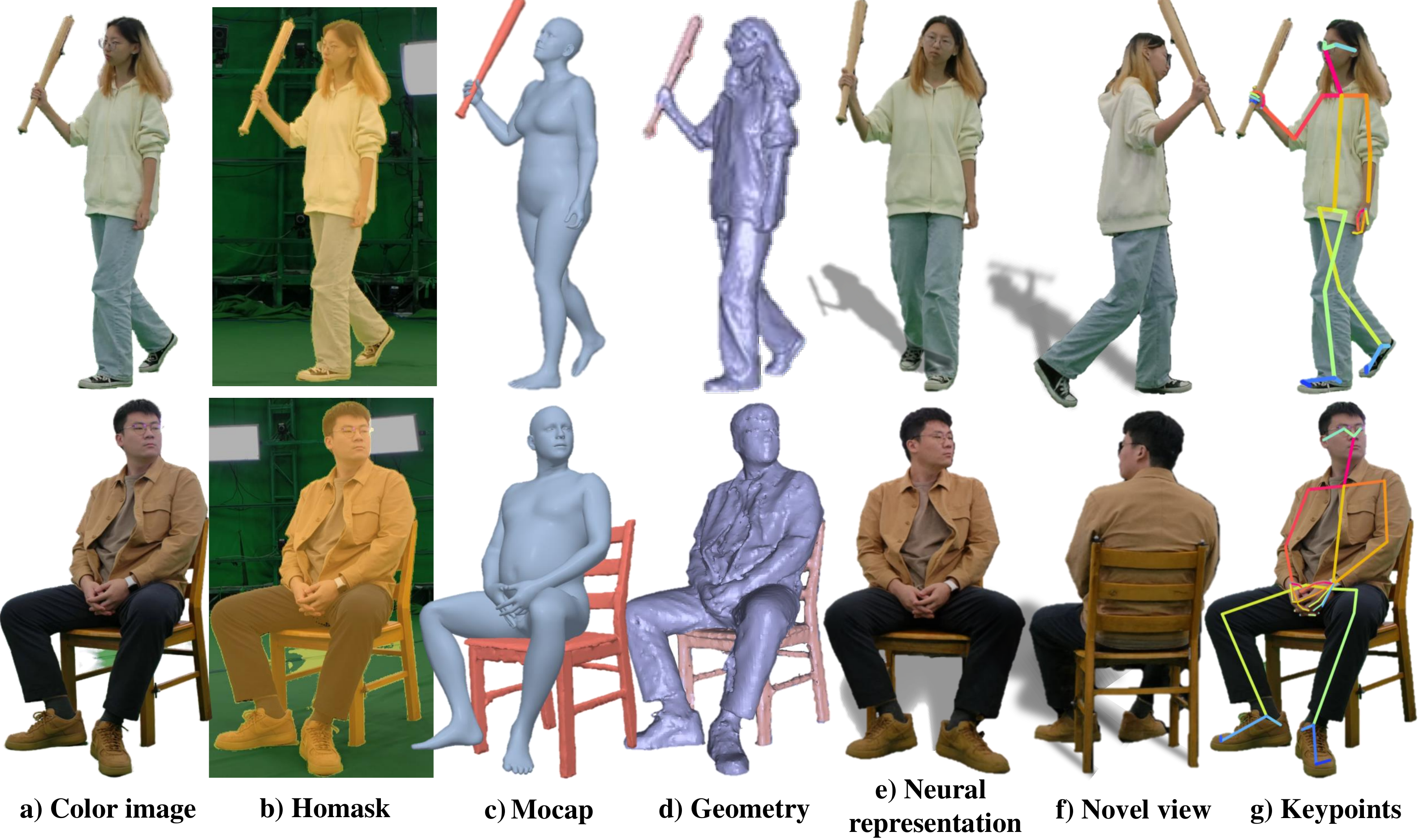}
\caption{\textbf{HODome Modality.} HODome features multiple modalities of data format and annotations, including a) Color image, b) Human-object mask (Homask), c) MoCap (SMPL-X parameters and located object template), d) Geometry, e) digital assets in neural representations, f) Novel view and g) Keypoints (25 for bodies and 42 for hands with human annotation).}
\label{Overview}
\end{figure}

\begin{table*}[t]\scriptsize
\centering
\renewcommand\arraystretch{1}
\resizebox{2\columnwidth}{!}{
\begin{tabular}{l|l|l|l|l|l|l|l|l|l|l|l|l} 
\toprule[2pt]
\makecell{Datasets} & \makecell{\#Cam-view} & \makecell{\# Frame(M)} & \makecell{Resolution} & \makecell{Fps} & \makecell{Marker}  & \makecell{ Obj. \\ Num.}   & \makecell{Human \\ Annot.}  & \makecell{Novel views}  & \makecell{Geometry}& \makecell{ Human \\Textured} & \makecell{Neural\\ Representation}  & \makecell{ Object \\ Appearance} \\
\cline{1-13}
\makecell{NTU~\cite{liu2019ntu}}     &  \makecell{3} &  \makecell{$\sim$ 34} & \makecell{1920 $\times$ 1080} &  \makecell{30} & \makecell{ $\times$}&   \makecell{NA}&  \makecell{\Checkmark}  & \makecell{NA}& \makecell{$\times$}& \makecell{$\times$}& \makecell{$\times$} & \makecell{$\times$}\\ 
\makecell{PiGr~\cite{savva2016pigraphs}}    &  \makecell{1} &  \makecell{0.1}&  \makecell{960 $\times$ 540} &  \makecell{5} &  \makecell{$\times$}  & \makecell{NA} &  \makecell{\Checkmark}  & \makecell{NA}& \makecell{$\times$}  & \makecell{$\times$}& \makecell{$\times$} & \makecell{\Checkmark}\\ 
\makecell{GRAB~\cite{taheri2020grab}}     &  \makecell{$\times$} &  \makecell{NA}&  \makecell{NA}&  \makecell{NA} &  \makecell{\Checkmark} & \makecell{51} & \makecell{$\times$}   & \makecell{$\times$}& \makecell{$\times$} & \makecell{$\times$}& \makecell{$\times$} & \makecell{$\times$}\\ 
\makecell{PROX~\cite{hassan2019resolving}}     &  \makecell{1}&  \makecell{0.1}&  \makecell{1920 $\times$ 1080}&  \makecell{30} & \makecell{\Checkmark}  & \makecell{NA} &\makecell{\Checkmark}    &\makecell{$\times$} &\makecell{$\times$}  & \makecell{$\times$}& \makecell{$\times$} & \makecell{\Checkmark}\\ 
\makecell{BEHAVE~\cite{bhatnagar2022behave}}     &  \makecell{4} &  \makecell{0.15}&  \makecell{2048 $\times$ 1536}&  \makecell{30} & \makecell{$\times$}  & \makecell{20} &  \makecell{$\times$}  &\makecell{$\times$} &  \makecell{$\times$}  & \makecell{$\times$}& \makecell{$\times$} & \makecell{\Checkmark}\\
\makecell{InterCap~\cite{huang2022intercap}}    &  \makecell{6} &  \makecell{0.07} &  \makecell{1920 $\times$ 1080} &  \makecell{30} & \makecell{$\times$}  & \makecell{10} & \makecell{\Checkmark}   &\makecell{$\times$} & \makecell{$\times$} & \makecell{$\times$}& \makecell{$\times$}& \makecell{\Checkmark}\\ 
\makecell{Our}  &  \makecell{76} &  \makecell{75} &  \makecell{3840$\times$ 2160} &  \makecell{60}  & \makecell{\Checkmark}  &\makecell{23}   & \makecell{\Checkmark}    & \makecell{\Checkmark} &   \makecell{\Checkmark} &   \makecell{\Checkmark}& \makecell{\Checkmark} & \makecell{\Checkmark} \\
\bottomrule[2pt]
\end{tabular} }
\caption{$\textbf{Dataset Comparisons}$. We compare our proposed HODome dataset with the existing publicly available human-object dataset. HODome has the largest scale of human-object interactions in terms of the number of frames (\#Frame), camera view (\#Cam-view), and modality. ``Obj. Num.'' represents the object number. ``Human Annot.'' represents annotation from professional annotators.}
\label{Dataset Comparion}
\end{table*}





\subsection{Neural Human Rendering}

%
Various 3D data representations have been explored for neural human rendering, such as point-clouds~\cite{aliev2020neural, wu2020multi, pang2021fewshot}, textured meshes~\cite{liu2020neuralhuman, liu2019neural, shysheya2019textured, habermann2021real}, and volumes~\cite{lombardi2019neural, lombardi2021mixture}.  
%
Implicit occupancy function-based methods~\cite{saito2019pifu, saito2020pifuhd, huang2020arch, he2021arch++} can recover detailed 3D human geometry from sparse 2D images without faithful appearance synthesis.
%
The recent NeRF~\cite{mildenhall2020nerf} technique brings huge potential for 3D photo-realistic view synthesis~\cite{wang2021ibrnet,chen2021mvsnerf, meng2021gnerf, 9466273, yu2021plenoctrees, Niemeyer2021Regnerf, liu2022neuray, mueller2022instant, verbin2022ref, chen2022tensorf} and geometry modeling~\cite{wang2021neus, long2022sparseneus, Cai2022NDR}. 
Further explorations extend it to dynamic scenes~\cite{STNeRF_SIGGRAPH2021, wang2022fourier, pumarola2020d, park2021hypernerf, 10.1145/3528223.3530086, Cai2022NDR}, especially for humans. Existing works equip NeRF with pose-embeddings~\cite{liu2021neuralActor, zhao2022humannerf, kwon2021neural, noguchi2021neural, peng2021neural}, learnable skinning weights~\cite{peng2021animatable, li2022tava, wang2022arah} and even generalization across individuals~\cite{zhao2022humannerf, wang2021ibutter, kwon2021neural}. However, these works only focus on a single person. 
The most relevant works~\cite{STNeRF_SIGGRAPH2021, 10.1145/3528233.3530704} aim to model multi-person interactions, but they cannot handle more complex interactions due to the lack of large-scale human-object interactions dataset. 


\subsection{Human-object Modeling}


Only a few works~\cite{hassan2019resolving, taheri2020grab, bhatnagar2022behave, huang2022intercap, xie2022chore, zhang2020perceiving} consider to jointly model whole-body interactions. Early solutions~\cite{collet2015high,schoenberger2016sfm,motion2fusion} that reconstruct meshes with per-frame texture maps, which are vulnerable to occlusions and lack of textures. 
Recently, several  works~\cite{hassan2019resolving,taheri2020grab,bhatnagar2022behave,huang2022intercap} explore the relationship via several interaction constraints, such as contact map~\cite{taheri2020grab,bhatnagar2022behave,huang2022intercap}, spatial arrangement~\cite{zhang2020perceiving} and physically plausible constraint~\cite{yi2022human}. However, these methods only produce a relative arrangement. The most relevant works~\cite{RobustFusion,sun2021neural,jiang2022neuralhofusion} aim to render human-object interactions with volumetric fusion \cite{RobustFusion}, neural texturing blending\cite{sun2021neural} and volumetric rendering \cite{jiang2022neuralhofusion}. However, without sufficient appearance supervision, their quality is limited. Comparably, our layer-wise neural representations can be converted to high-quality geometry and support separate free-view rendering for human or object.

\begin{figure*}[t!]
  \centering
  \includegraphics[width=1.0\linewidth]{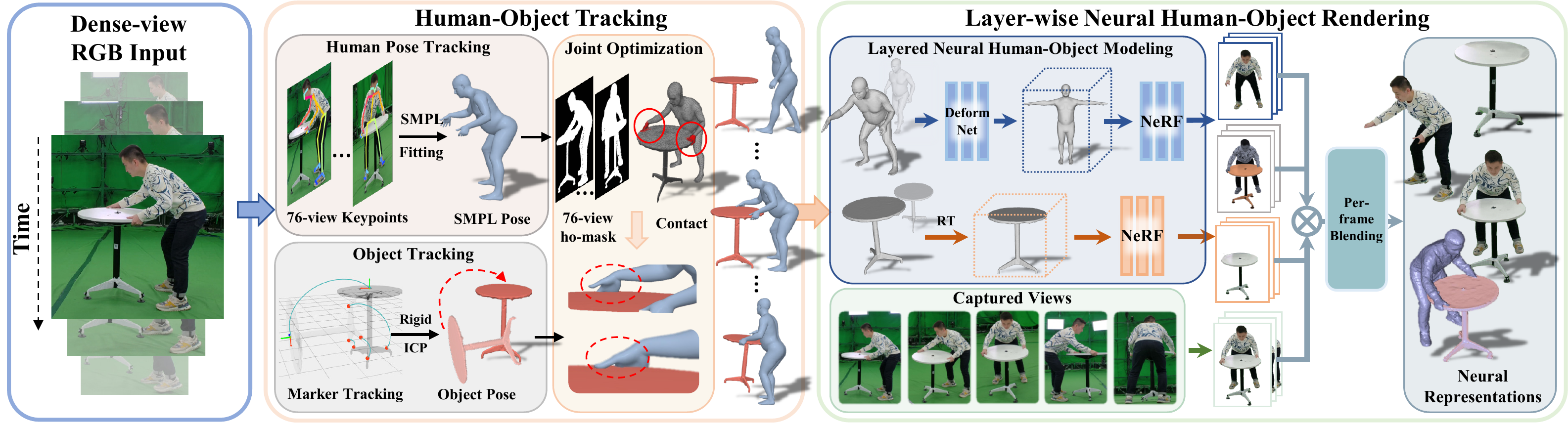}
 \caption{\textbf{Overview of NeuralDome.} Given the 76-view RGB stream as input, we first jointly track human skeletal motion and object rigid motion. Then with the tracked motion priors, we decouple the human-object interaction scenes via a layer-wise neural rendering scheme, to generate human/object renderings separately and corresponding acceptable segmentation maps. Blending with the high-fidelity captured views, we obtain the layer-wise neural representations in our HODome dataset.}
  \label{fig:pipeline}
\end{figure*}


\subsection{Human-centric Dataset}

A variety of human-centric datasets have been developed for human-only capturing or rendering tasks. Early datasets combine multi-camera RGB video capture with synchronized ground-truth 3D skeletons~\cite{h36m_pami,sigal2010humaneva,Joo_2017_TPAMI}, parametric model~\cite{Lassner:UP:2017,von2018recovering}, scanned mesh~\cite{Zhang_2017_CVPR,hasler2009statistical}, without human-object interaction. Recent datasets capture human-object interactions using optical markers~\cite{taheri2020grab} or sparse RGB sensors~\cite{liu2019ntu,bhatnagar2022behave}, without sufficient appearance supervision. As a result, the literature on neural human-object rendering~\cite{sun2021HOIFVV,jiang2022neuralhofusion} is surprisingly rare. High-end works~\cite{collet2015high, guo2019relightables} use dense cameras for reconstruction and rendering of humans and objects through mesh reconstruction and motion tracking, but without texture and neural representations. To fill this gap, we propose NeuralDome for capturing and rendering human-object interactions, facilitating various generative human-object tasks.

\section{HODome Dataset}




We present the HODome dataset for capturing and rendering photo-realistic human-object interactions. It consists of 274 human-object interacting sequences, covering 23 diverse 3D objects and 10 subjects (5 males and 5 females) under various apparels. For each sequence, we record the naturally interacting scene for about 60s using 76 RGB cameras for dense-view at 3840 $\times$ 2160 resolution and 60 frame-per-seconds (Fps), resulting in roughly 75 million frames.
HODome consists of rich labels covering different aspects of HOI capturing and rendering labels (see Fig.\ref{Overview}).
See Tab.~\ref{Dataset Comparion} for comparison with other datasets.

\subsection{Data Capturing System}
To construct the HODome dataset, we use 76 Z-CAM cinema cameras with sufficient appearance supervision for neural rendering tasks and 16 Optitrack MoCap cameras~\cite{Optitrack} for accurate human-object tracking tasks. The Z-CAM system and Optitrack system are synchronized to record the RGB and MoCap data together. We use a publicly available tool~\cite{realitycapture} to estimate the intrinsic camera parameters and extrinsic camera parameters.

\subsection{Dataset Modality}


\myparagraph{Human MoCap and Object 6D Pose.} 
For appearance realism, we do not place markers on human actors. Thus we detect the human joints from 76 RGB images by running the whole-body Openpose\cite{cao2017realtime} to perform markerless MoCap. And we treat each object as rigid and solve the rigid object pose estimation from markers using the Iterative Closest Point(ICP) algorithm~\cite{besl1992method,zhou2018open3d}. See Fig. \ref{fig:pipeline} for the pipeline of our method.


\myparagraph{Human-Object Neural Representations.} 
Under the dense-view setting, one can apply off-the-shell neural techniques, i.e., Instant-NSR~\cite{zhao2022human} and Instant-NGP~\cite{mueller2022instant} to efficiently obtain per-frame geometry and novel-view synthesis of the whole human-object sequence, respectively. However, neither schemes separate object from human subject. We therefore propose a layer-wise neural scheme to separately recover neural representations of human and object, enabling both high-fidelity geometry reconstruction and free-view appearance rendering (Sec.~\ref{Sec: Neural HOI}).



\myparagraph{Data Annotation.} Our dataset includes 23 objects with varying scales and interaction types. Each object was pre-scanned using off-the-shelf multi-view software packages~\cite{realitycapture}. Further, following the previous methods~\cite{bhatnagar2022behave, taheri2020grab}, we annotated its pseudo contact label that computes from a threshold distance. To provide more accurate results for the quantitative benchmark, we annotate a separate \emph{quantitative subset} as our test set with human-annotated segmentation and hand joints by the professional annotator. Our datasets will be available for research purposes.

\section{Neural Modeling on HODome}  
%
We introduce a neural pipeline to produce the rich digital assets from the dense-view input of each human-object interaction (HOI) sequence in HODome, including accurate tracking, high-quality geometry reconstruction and novel-view rendering. 
As illustrated in Fig.~\ref{fig:pipeline}, given 76-view videos, we first perform a joint optimization scheme to accurately capture both the human skeletal motions and object rigid motion (Sec.~\ref{sec: hocapture}).  
Then, based on the tracked human-object motions and shape priors, we decouple the humans and objects in HOI scenarios via a layer-wise neural human-object rendering scheme and a corresponding HOI-aware optimization strategy (Sec.~\ref{Sec: Neural HOI}). 
%
%
Such digital assets of humans, objects and the entire HOI scenes can naturally enable further geometry and appearance analysis of HOI scenarios. 

\subsection{Human-object Tracking} \label{sec: hocapture}

Here we introduce our human-object tracking scheme with the aid of object markers.

\myparagraph{Tracking initialization.} 
We initialize the human tracking process by fitting the SMPL-X model to the keypoints of each view. Note that we utilize the off-the-shell toolbox Easymocap~\cite{easymocap} to get the initial per-frame parameters, i.e., pose $\theta_t$, shape $\beta_t$, facial expression $\psi_t$, and translation $\gamma_t$ on each frame $t$.
%
As for the object, following GRAB~\cite{taheri2020grab}, 
we regard each object as rigid and only estimate the rotation $R_t \in \mathbb{SO}(3)$ and rigid translation $T_t \in \mathbb{R}^{3}$ with respect to its pre-scanned template.
Specifically, we conduct the rigid-ICP technique~\cite{besl1992method,zhou2018open3d} to compute the rigid transformation between the per-frame markers, which serves as the initial object pose. 

\myparagraph{Joint optimization for human-object tracking.} 
With human and object pose initialization, we further perform a joint optimization scheme to ensure correct human-object contact. 
We impose constraints to ensure plausible interactions as the following form: 
\begin{equation} \label{eq:tracking}
\begin{aligned}
E(\beta_t,\theta_t,\psi_t, \gamma_t, R_t,T_t) &= E_{\text{smpl}} + \lambda_{\text{contact}} E_{\text{contact}} +  \\  &  \lambda_{\text{homask}} E_{\text{homask}} + \lambda_{\text{maker}} E_{\text{maker}},
\end{aligned}
\end{equation}
where $E_{\text{smpl}}$ is the multi-view data fitting term describing the $\ell_2$ distance between estimated joints and detected joints, following SMPLify-X~\cite{pavlakos2019expressive} and PROX~\cite{hassan2019resolving}.
Besides, we impose contact term $E_{\text{contact}}$, human-object silhouette loss term $E_{\text{homask}}$, and object marker align term $E_{\text{maker}}$ to ensure the plausibility of human-object interaction tracking. 
Due to page limitation, we have to defer more details of these terms in the Appendix.



\subsection{Layer-wise Neural Human-Object Rendering} \label{Sec: Neural HOI}

\begin{figure}[t]
  \centering
  \includegraphics[width=1.0\linewidth]{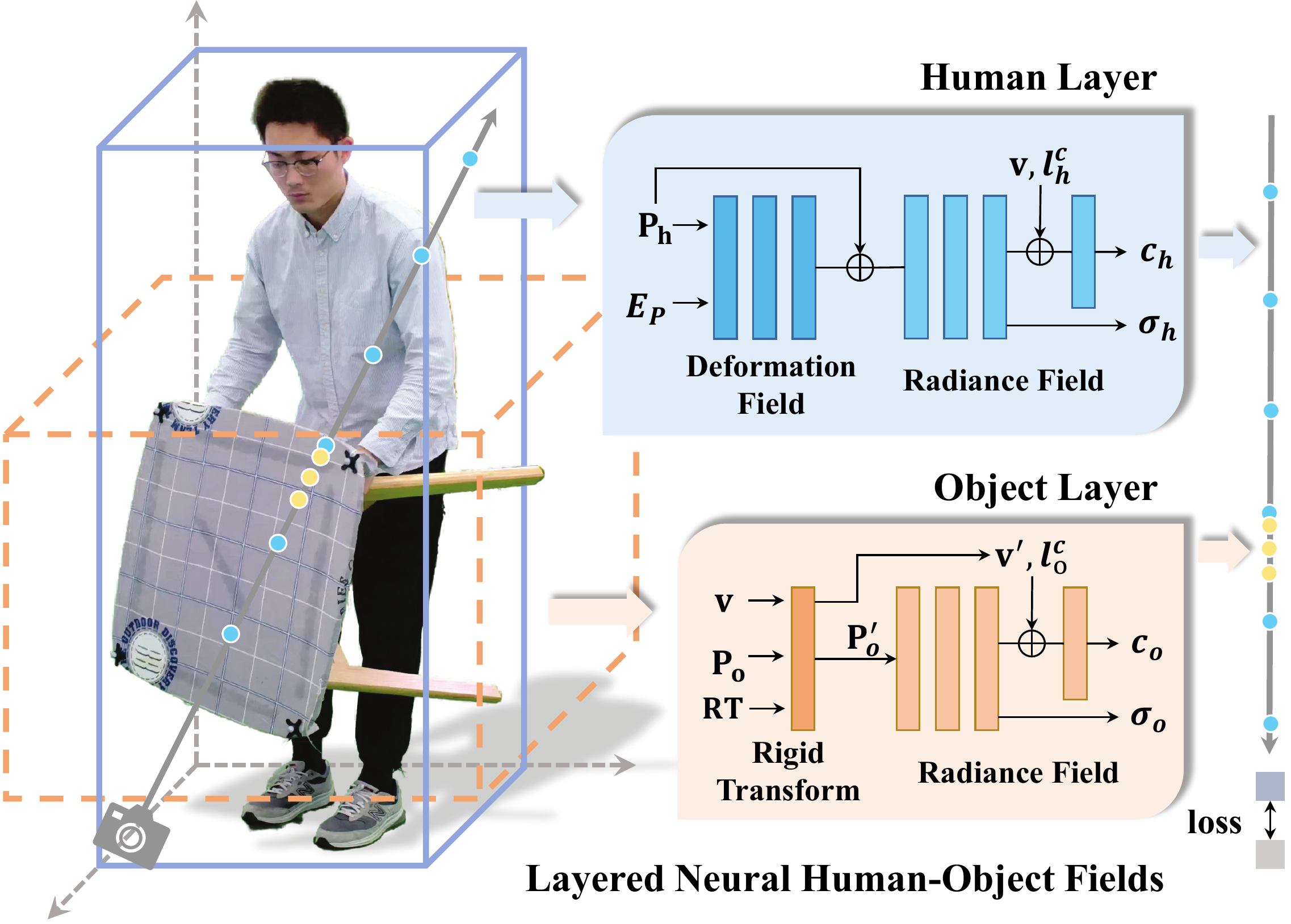}
  \caption{\textbf{Layer-wise Neural Human-Object Rendering Details.} We sample a ray uniformly in the bounding box of the performer and near the template for the object. We condition both the human and object layer with appearance codes ($l_o^c, l_h^c$). The dynamic human layer contains an additional pose embedding ($E_p$) conditioned deformation field. The sample colors and densities are merged, sorted and accumulated into pixel colors.} 
  \label{fig:layer_representation}
\end{figure}

%
%
Here we introduce our layer-wise image generation pipeline via neural rendering, without tedious manual efforts for geometry separation. 

%

\myparagraph{Layered neural human-object modeling.} 
For ease of instances separation, we represent the human-object interaction (HOI) scenes as a continuous layered neural radiance field, following ST-NeRF~\cite{STNeRF_SIGGRAPH2021}, as shown in Fig.~\ref{fig:layer_representation}.
%
To leverage the shape prior imposed by the captured parametric SMPL-X model, we adopt a pose embedded dynamic NeRF similar to HumanNeRF~\cite{zhao2022humannerf} as the human layer. 
We use the skeletal pose to bridge the live frames to canonical space and an additional deformation MLP to learn subtle non-rigid deformation. 
Rather than using the pixel-aligned image feature, we leverage latent codes for time-varying human appearance capture to get rid of input images.
Opposite to the human layer, we model objects as rigid static radiance fields in canonical space. 
The live frames are transformed to canonical space via object poses to maintain a globally consistent density field.
Similarly, we adopt appearance latent codes for the time-varying shadow during interactions.


\begin{figure*}[t!]
  \centering
   \includegraphics[scale = 0.5]{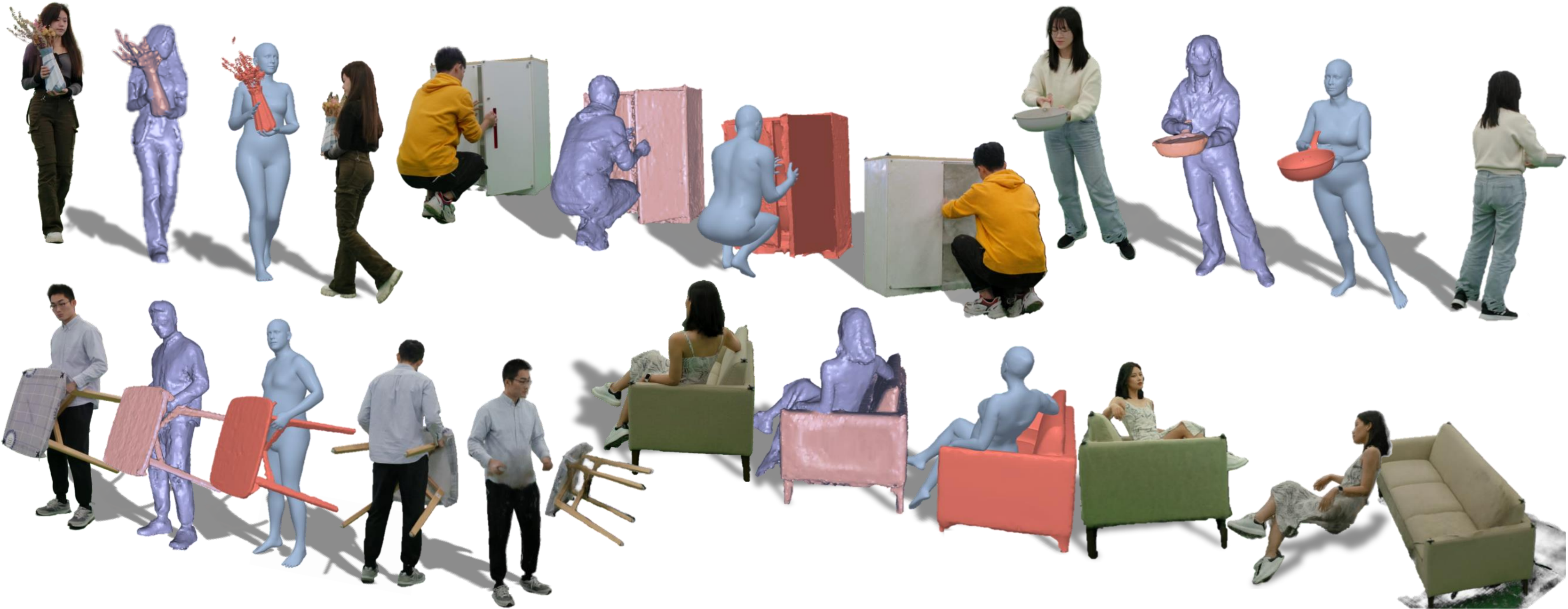}
 \caption{\textbf{The geometry, MoCap and neural modeling results from our dataset}. HODome includes various interaction sequences, such as ``holding a vase'', ``opening the cupboard'', ``sitting on the sofa'' and ``moving the table''.}
  \label{fig:gallery}
\end{figure*}

\myparagraph{Dynamic human-object volume rendering. }
We utilize the layered volume rendering technique described in ST-NeRF~\cite{STNeRF_SIGGRAPH2021} to render our neural human-object scenes.
For a camera ray intersecting with the $i_{th}$ entity at any timestamp, we compute the ray segment as the depth of intersection points $d_f^j$ and $d_f^i$. 
%
We evenly partition each segment into $N$ bins and sample one point uniformly from each bin:

\begin{small}
    \begin{equation}
        p_j^i \! \sim \! \mathcal{U} \left[ d_n^i \!+\! \frac{j\!-\!1}{N}(d_f^i \!-\! d_n^i), d_n^i \!+\! \frac{j}{N}(d_f^i \!-\! d_n^i)\right], j \! \in \! [1, 2, ..., N], 
    \end{equation}
\end{small} 
where $\mathcal{U}$ means uniform distribution.
We impose the shape prior encoded in the tracked object template, on the ray sampling scheme for efficient training. We compute the ray-object intersection and uniformly sample a few points in the narrow segment where the first intersection point lies.
For human, we simply sample in the ray-bounding box intersection segment.
The samples from different segments are then merged and sorted by depth into $M$ samples in total.
We then compute a pixel's color by accumulating the radiance at sampled points:

\begin{equation}
    C = \sum_{i=1}^{M} T(p_i)[(1 - e^{-\sigma_{p_i}\delta_{p_i}})c_{p_i}],
\end{equation}
where $\delta_{p_i}$ is the distance between adjacent points, $c_{p_i}$ and $\sigma_{p_i}$ are color and density, and $T(p_i) = \prod_j^{i-1}e^{-\sigma_{p_j}\delta_{p_j}}$.

\myparagraph{HOI-aware training scheme.} 
Here we introduce an effective optimization scheme to train our neural layer-wise HOI model. 
We first leverage a photometric loss:
\begin{equation}
    \mathcal{L}_{c} = \sum_{\mathbf{r}\in\mathcal{R}}\|C_\mathbf{r} - \hat{C}_{\mathbf{r}}\|_2^2,
\end{equation}
where $\mathbf{r}$ is a ray in the training ray set $\mathcal{R}$, $C_\mathbf{r}$ and $\hat{C}_\mathbf{r}$ are the corresponding rendered and observed colors.  

Note that in HOI scenarios, the human layer inevitably intersects with other object layers. This fact introduces ambiguities and cannot be addressed by a simple photometric loss. 
Thus we design three additional regularizers based on the tracked object template to alleviate this issue.
To enforce the object layers to be solid surface radiance fields that do not contribute to human appearance, we adopt an occupancy and a sparsity regularizer inside and outside the object template, i.e.,
\begin{equation}
    \mathcal{L}_{o} = \sum_{p\in \mathcal{O}}(\Omega_{\!-\!}(p, \mathcal{M})\| e^{ (-\sigma_{p}) }\|_2^2 + \Omega_{\!+\!}(p, \mathcal{M})\|\sigma_{p}\|_2^2),
\end{equation}
where $\mathcal{M}$ is the template mesh, $\mathcal{O}$ is a set of points randomly sampled in the object's bounding box, $\Omega_{\!+\!}$ and $\Omega_{\!-\!}$ indicate whether a point locates outside or inside the mesh.

Recall that we have encoded human-object contact prior by jointly tracking human-object at the skeletal-pose level (Eq.~\ref{eq:tracking}).
However, due to the misalignment of SMPL shape and actual human geometry, the human body and clothes may still overlap the object space, which goes against real-world physics.
Therefore, we implicitly constrain the deformation net to predict contact-aware non-rigid deformation:
\begin{equation}
    \mathcal{L}_{h} = \sum_{p\in\mathcal{H}} \Omega_{\!-\!}(p, \mathcal{M}) \|\sigma_{p}\|_2^2,
\end{equation}
where $\mathcal{H}$ is the set of random samples in the human's bounding box.

We further explore a weakly-supervising scheme that utilizes object texture cues to help decouple the human and object entities.
The key observation is that, with accurate object tracking, the object radiance field fused across frames can be rendered with higher confidence. 
That is, we train the entire scene first and render the objects only to capture views, and the pixels similar to the captured ones are labeled as object pixels.
In this way, we obtain coarse object segmentation maps $\mathcal{S}$ that serve as pseudo supervision:
%
we adopt label-wise integration~\cite{10.1145/3528233.3530704} to render the layer labels $\mathbf{s}$ of rays:
\begin{equation}
    \mathbf{s} = \sum_{i=1}^{M} T(p_i)[(1 - e^{-\sigma_{p_i}\delta_{p_i}})\mathbf{l}_{p_i}], 
\end{equation}
where $\mathbf{l}_{p_i}$ is one-hot label indicates which layer the point $p_i$ belongs to. 
We apply a semantic loss for object rays $\mathcal{S}_o$:
\begin{equation}
    \mathcal{L}_s = \sum_{\mathbf{r}\in\mathcal{S}_o}\|\mathbf{s}_{\mathbf{r}} - \mathbf{\hat{\mathbf{s}}_{\mathbf{r}}}\|_2^2,
\end{equation}
where $\mathbf{\hat{\mathbf{s}}_{\mathbf{r}}}$ is the pseudo semantic label of ray $\mathbf{r}$. Note that additional loss can be applied if accurate labels are provided.

\myparagraph{Layer-wise neural representation in HODome.}
To further use the data, we can render reliable segmentation maps and occlusion-free appearances for humans and objects respectively. 
However, the renderings tend to be blurred since the information of all the frames is fused.
Hence we introduce an enhancement scheme to construct separate high-quality per-frame neural representation for human/object:
we blend the visible regions of the high-fidelity observed input views to rendered human/object images using the segmentation maps.
The Instant-NGP~\cite{mueller2022instant} and Instant-NSR~\cite{zhao2022human} techniques are then applied to reconstruct separate human/object in tens of seconds.
Finally, we obtain layer-wise representation which enables real-time photo-realistic rendering and benefits the thorough analysis of HOI scenarios. 

\myparagraph{Implementations.}
To build our neural representations, we train our layered models using Adam optimizer with a learning rate that starts from $5e^{-4}$ and decays exponentially. 
Using the 76-view 4K resolution videos, the training process takes 8-10 hours on a single NVIDIA 3090 GPU for acceptable appearance and segmentation rendering. 
Thanks to Instant-NGP~\cite{mueller2022instant} and Instant-NSR~\cite{zhao2022human}, the layer-wise neural modeling process is conducted in seconds per frame.

\section{Experiments}

In this section, we compare our neural pipeline with existing state-of-the-arts. We first demonstrate the effectiveness and the generalization ability of our neural modeling approach (\ref{Analysis}). Next, we implement NeuralDome on three specific tasks (\ref{benchmark}).




\subsection{Analysis of Neural Rendering } \label{Analysis}

\begin{figure}[t]
  \centering
  \includegraphics[width=1.0\linewidth]{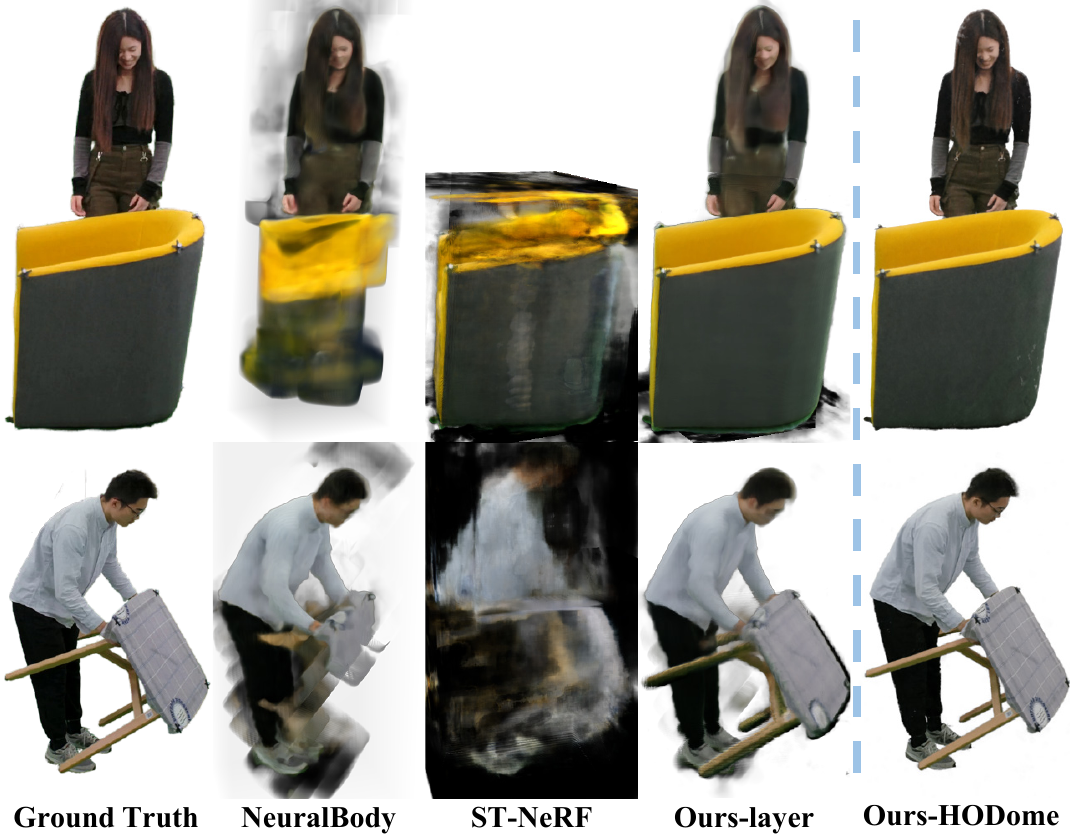}
  \caption{\textbf{Qualitative comparisons} of novel view synthesis results. We show ground truth and synthesized images of novel view for NeuralBody~\cite{peng2021neural}, and ST-NeRF~\cite{STNeRF_SIGGRAPH2021} and our layered human-object representation. Our approach achieves the best performance. We further illustrate our high-fidelity neural representations in HODome in the last column.}
  \label{fig:nr_comparison}
\end{figure}


\begin{table}[t]
\centering
\renewcommand\arraystretch{1.1}
\resizebox{1\linewidth}{!}{
\begin{tabular}
{llcccccc}
\toprule[2pt]
\multicolumn{2}{l}{Methods} & \multicolumn{2}{c}{NB~\cite{peng2021neural}} & \multicolumn{2}{c}{ST-NeRF~\cite{STNeRF_SIGGRAPH2021}} & \multicolumn{2}{c}{Ours}  \\
\cmidrule(r){1-2} \cmidrule(r){3-4} \cmidrule(r){5-6} \cmidrule(r){7-8}
\multicolumn{2}{l}{Scenes} & PSNR$\uparrow$ & SSIM$\uparrow$ & PSNR$\uparrow$ & SSIM$\uparrow$ & PSNR$\uparrow$ & SSIM$\uparrow$  \\
\cline{1-8}
Bigsofa   &  &  19.33  &  0.896  &  24.02  &  0.886  &  \textbf{32.28}  &  \textbf{0.958}   \\ 
Sofa      &  &  26.73  &  0.965  &  28.49  &  0.958  &  \textbf{35.62}  &  \textbf{0.987}   \\
Table     &  &  21.94  &  0.933  &  22.44  &  0.900  &  \textbf{27.88}  &  \textbf{0.949}   \\ 
\cline{1-8}
Average   &  &  22.67  &  0.931  &  24.99  &  0.915  &  \textbf{31.93}  &  \textbf{0.964}   \\ 
\bottomrule[2pt]
\end{tabular}
}
\caption{\textbf{Quantitative comparisons} of novel view synthesized appearance on different human-object interaction sequences.}
\label{tab: nr_comarison}
\end{table}



\myparagraph{Comparisons.}
Here we compare our layer-wise neural human-object rendering approach, denoted as 'Ours-layer', with recent state-of-the-art methods, i.e., ST-NeRF~\cite{STNeRF_SIGGRAPH2021}, NeuralBody(NB)~\cite{peng2021neural}, in human-object interaction scenario. For fair comparisons, we select 20 cameras surrounding the performer for training and others for evaluation. We remove the background layer and train ST-NeRF without the semantic label as ours. Fig.~\ref{fig:nr_comparison} shows several appearance synthesis results. NeuralBody has no ability to model objects which leads to artifacts in the density field. ST-NeRF fails as the bounding boxes of the performer and object almost completely overlap. Our method achieves both human and object modeling and rendering and thus further enables generating high-fidelity neural representation denoted as 'Ours-HODome'. Tab.~\ref{tab: nr_comarison} further illustrates our method significantly outperforms the baselines.

\begin{figure}[t]
  \centering
  \includegraphics[width=1.0\linewidth]{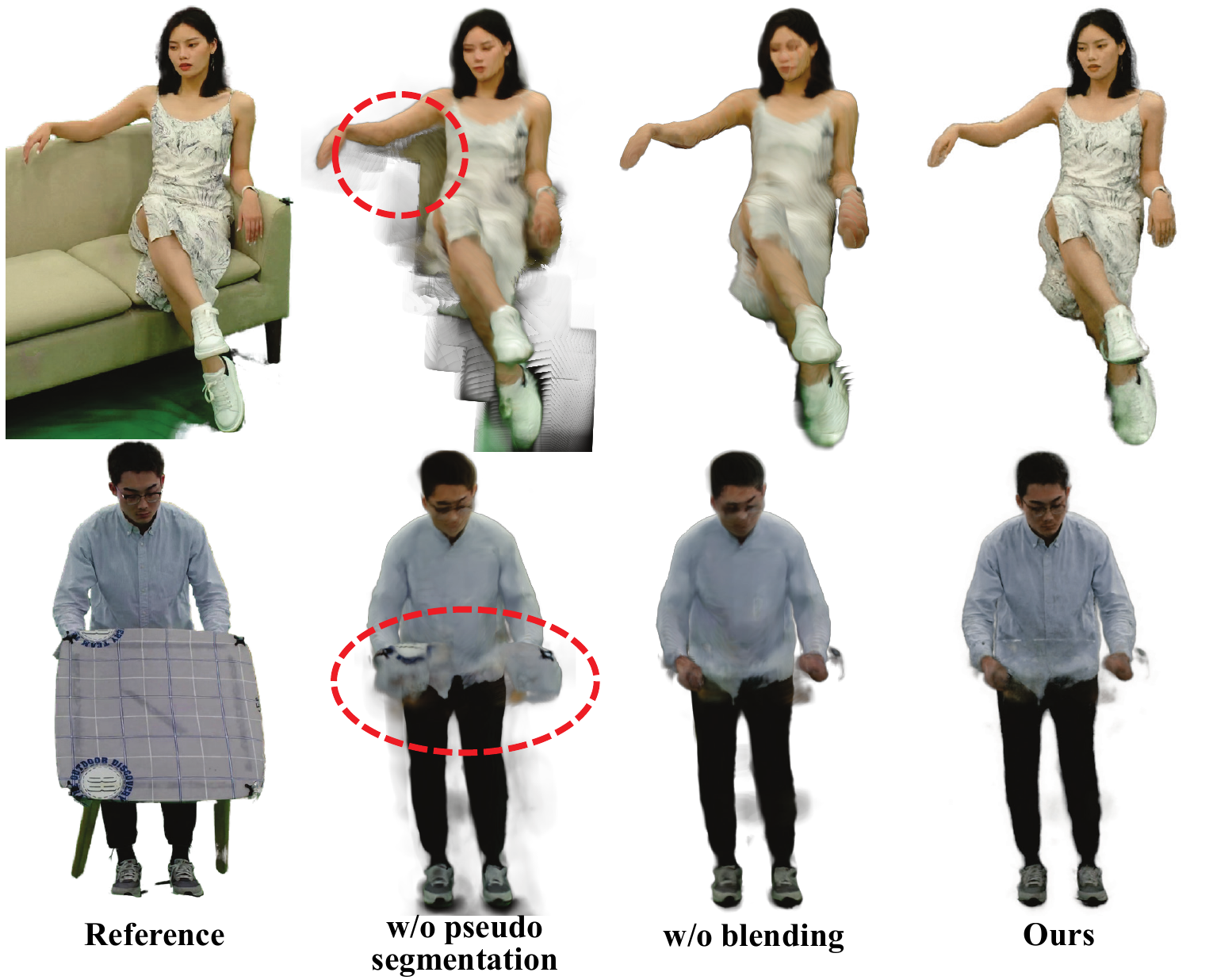}
  \caption{Qualitative evaluation of each strategy's contribution to layer-wise neural representation generation. The weakly-supervising scheme effectively helps decouple human and object and our blending scheme generates high-quality digital assets.}
  \label{fig:nr_ablation}
\end{figure}

\myparagraph{Evaluation.}
Here we further evaluate how our scheme contributes to the generated layer-wise neural representations.  Fig.~\ref{fig:nr_ablation} shows the single human layer where the performer is "sitting" in the air.  
Note that we have no ground truth of the specific human/object layer, thus we conduct qualitative evaluations only.  
Let \textbf{w/o pseudo segmentation} and \textbf{w/o blending} denote our human assets generated without pseudo semantic loss and blending-based enhancement scheme. 
It demonstrates that the pseudo semantic loss effectively helps decouple the human and object layer and our blending scheme further boosts the appearance fidelity of the blurred temporal-fused model to a photo-realistic level.
 
\subsection{Task and benchmark} \label{benchmark}

\noindent\textbf{Human-object Capture Benchmark.}
HODome provides multi-view sequences with synchronized ground truth of capturing. To demonstrate the capability, we provide a benchmark for human-object capture, shown in Tab.~\ref{Capture benchmark}. For detailed metric explanation please refer to \cite{bhatnagar2022behave, yi2022human, xie2022chore}. PHOSA~\cite{zhang2020perceiving} used the contact map from hand-crafted annotations, suffering from coarse contact information. To demonstrate the contact quality, we use pseudo labels from our pipeline as input for comparison. As shown in Tab.~\ref{Capture benchmark}, the pseudo contact label computed from our pseudo label helps PHOSA achieve better results. While CHORE\cite{xie2022chore} outperforms them with the distance fields predicted from neural networks. It shows the importance of a prepared human-object dataset that supports data-driven human-object capture model.

\begin{table}[t]
\centering
\renewcommand\arraystretch{1.1}
\renewcommand{\multirowsetup}{\centering}
\resizebox{1\linewidth}{!}{
\begin{tabular}{c|c|c|c|c|c|c} 
\toprule[2pt]
 {\multirow{2}{*}{\makecell{Method}}} & \multicolumn{4}{c|}{Separated evaluation} & \multicolumn{2}{c}{Joint evaluation} \\
\cline{2-7}  & \makecell{MPJPE$\downarrow$} & \makecell{\text{PA-MPJPE}$\downarrow$} & \makecell{Chamfer$_{h}\downarrow$} &\makecell{Chamfer$_{o}\downarrow$}& \makecell{V2V$\downarrow$} & \makecell{p.V2V$\downarrow$}  \\
\cline{1-7}
\makecell{Fit to input}    &  \makecell{16.11}  & \makecell{6.94}  & \makecell{86.66} &  \makecell{18.14} & \makecell{61.57}  &  \makecell{28.92}  \\
\makecell{PHOSA \cite{zhang2020perceiving}}   &  \makecell{14.88}  & \makecell{6.94} & \makecell{77.62}  & \makecell{16.64} & \makecell{54.59}  & \makecell{26.09}      \\ 
\makecell{CHORE\cite{xie2022chore}}   &    \makecell{10.21} &  \makecell{6.14} &  \makecell{86.58} & \makecell{7.69} & \makecell{44.93} &  \makecell{14.93}\\ 
\makecell{PHOSA w. cont.}    & \makecell{14.75}  & \makecell{6.90} & \makecell{76.95} & \makecell{16.61}  & \makecell{54.21}  & \makecell{25.63}   \\
\bottomrule[2pt]
\end{tabular}}
\caption{\textbf{Human object capture benchmark.} "Fit to input" represents the vanilla method that fits the object template to image and capture human with Frankmocap\cite{rong2021frankmocap}. 'PHOSA w. cont.' represents using the contact labels from our pseudo label.}
\label{Capture benchmark}
\end{table}

\begin{figure}[t!]
\includegraphics[scale = 0.35]{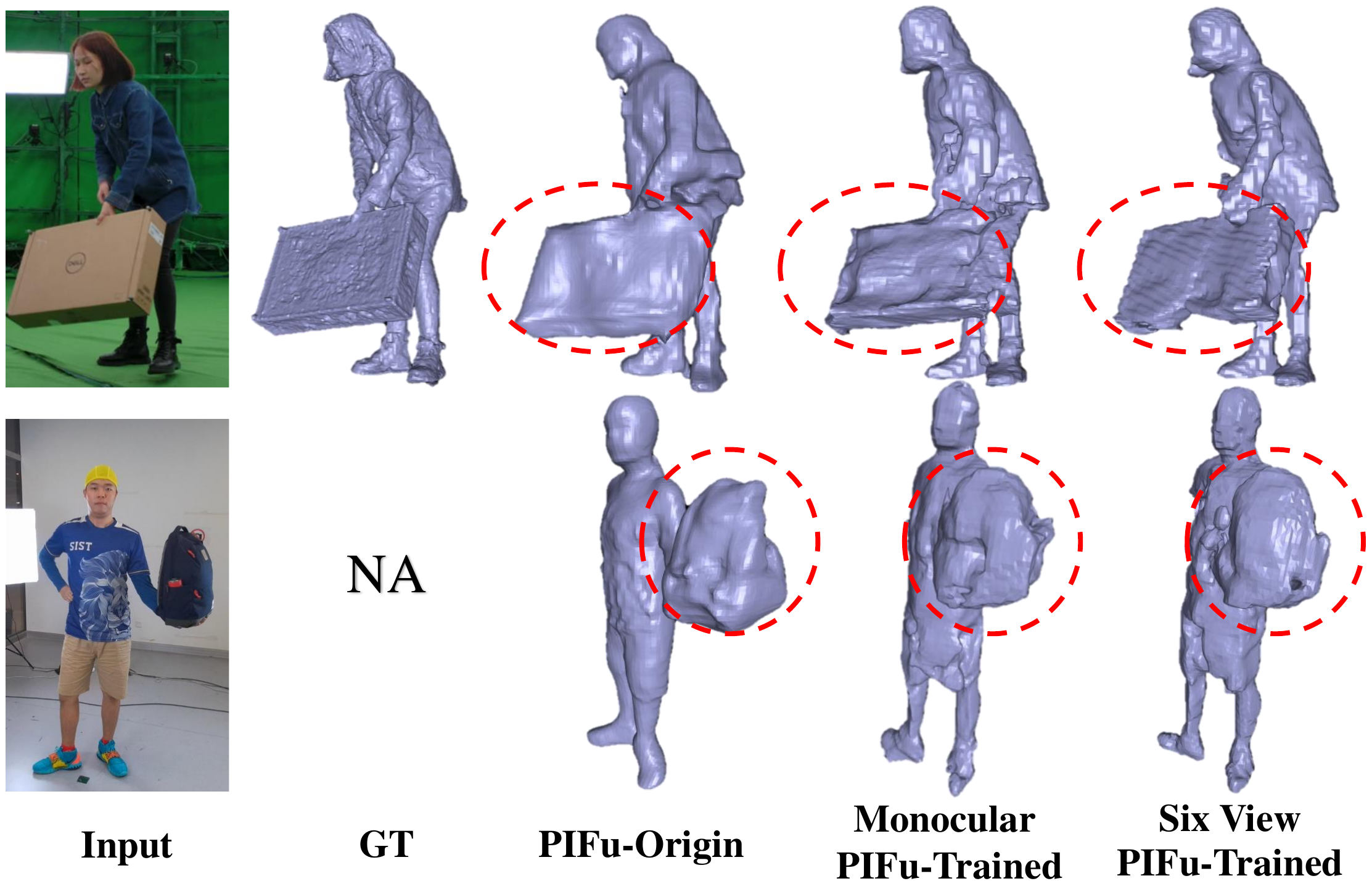}
\captionof{figure}{\textbf{Qualitative evaluation} on human-object reconstruction. We compare the original PIFu\cite{saito2019pifu} with trained PIFu on both images from our dataset and in-the-wild images. The original PIFu fails to predict the object depth, and PIFu trained on our dataset can approximately predict the shape of the object.}
\label{fig:geometry}
\end{figure}

\myparagraph{Human-object Geometry Reconstruction Benchmark.}
Benefiting from the dense-view setting, various human-only geometry reconstruction methods can step towards a human-object geometry reconstruction setting. We can benchmark the state-of-the-art algorithms in sparse view geometry reconstruction tasks. Hence, we evaluate PIFu \cite{saito2019pifu} by training on a subset of HODome. As shown in Fig. \ref{fig:geometry}, we compare the quality result between original PIFu and PIFu-trained on HODome and in-the-wild images. Compared to the original PIFu, our training set can improve the quality of the reconstructed shape of the objects.  For further quantitative analysis, we evaluate the performance using \textbf{P2S} and \textbf{CD} in Tab. \ref{geo table}. After training, the mesh prediction achieves higher accuracy. However, due to the limitation of PIFu, there is still much more that can be improved for human-object geometry reconstruction.

\begin{table}[t] \scriptsize
\centering
\renewcommand\arraystretch{1.1}
\begin{tabular}{l|c|c}
\toprule[1pt]
\makecell{Method} & \makecell{P2S $\times 10^{-4}$ $\downarrow$} & \makecell{Chamfer $\times 10^{-4}$ $\downarrow$} \\
\cline{1-3}
Origin PIFu\cite{saito2019pifu}    & 38.726   &  40.947       \\ 
Monocular PIFu-trained      &  14.653  & 14.483       \\ 
6-View PIFu-trained     &   \textbf{3.376}     & \textbf{4.901}     \\
\bottomrule[1pt]
\end{tabular}
\caption{\textbf{Geometry reconstruction benchmark}. Origin PIFu\cite{saito2019pifu} uses the pre-trained model to infer. Monocular PIFu is trained on a single view. 6-View PIFu is trained on 6-view inputs.}
\label{geo table}
\end{table}

\begin{figure}[t!]
\includegraphics[width=1.0 \linewidth]{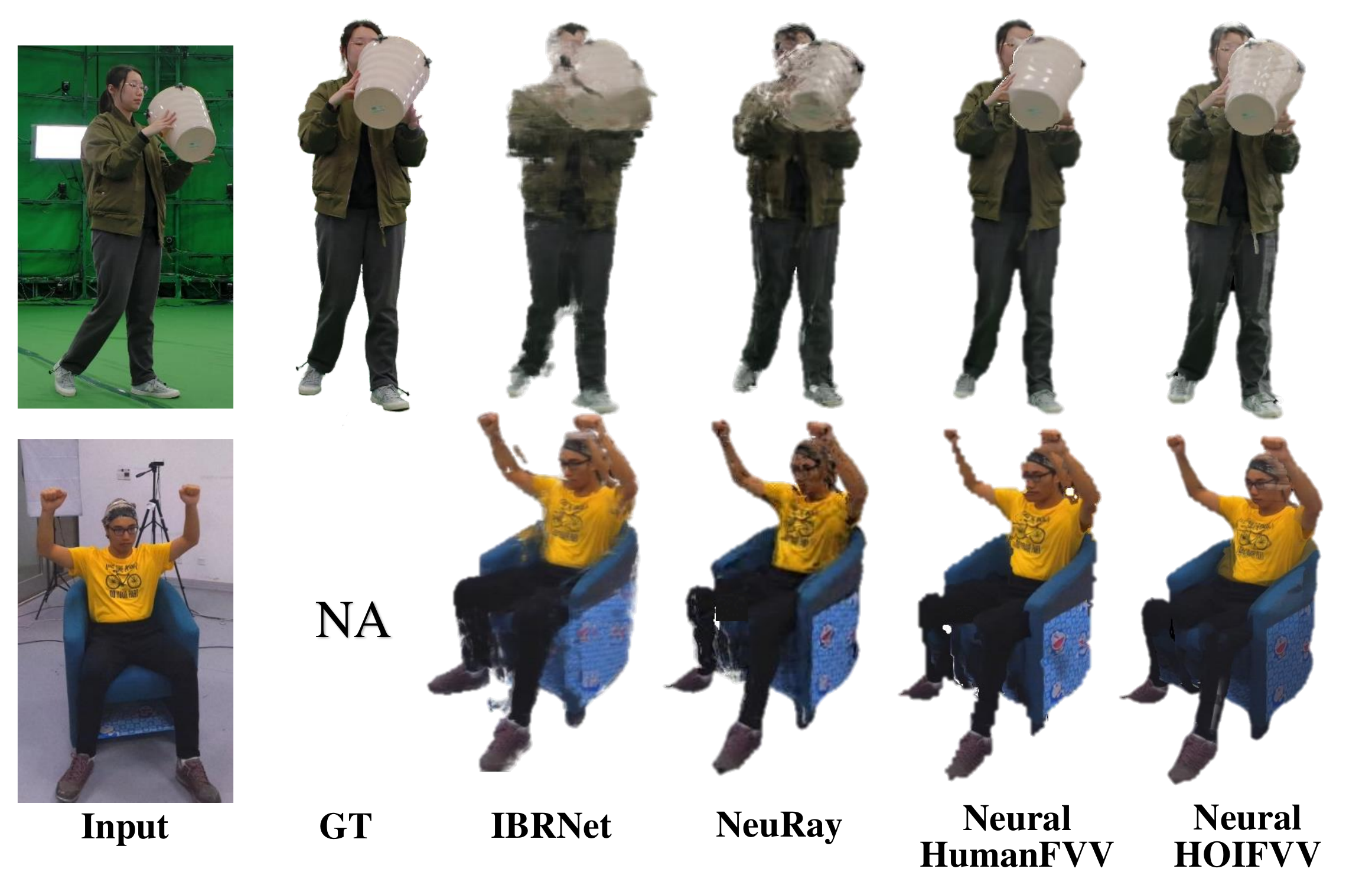}
\captionof{figure}{\textbf{Qualitative evaluation} on neural human-object rendering. We show the comparison of IBRNet\cite{wang2021ibrnet}, NeuRay\cite{liu2022neuray}, NeuralHumanFVV\cite{9578599} and our proposed NeuralHOIFVV. We render one novel view between two input camera views.}
\label{fig:render}
\end{figure}

\myparagraph{Sparse-View Human-object Rendering Benchmark.}
Our HODome supports various neural rendering tasks, even for human-object interactions. We provide a benchmark on sparse-view rendering tasks and evaluate on IBRNet \cite{wang2021ibrnet}, NeuRay \cite{liu2022neuray} and NeuralHumanFVV \cite{9578599}. Besides, we also provide a baseline, named NeuralHOIFVV, which uses the projection of trained PIFu's \cite{saito2019pifu} results as depth input, then applies the Neural Blending method presented in \cite{zhao2022human} to obtain novel-view synthesis. More details about NeuralHOIFVV are referred to the Appendix. We analyze the above methods on HODome and in-the-wild images. The qualitative and quantitative results are shown in Fig. \ref{fig:render} and Tab. \ref{tab: render}. Our proposed NeuralHOIFVV can obtain generative novel-view synthesis on HODome and in-the-wild inputs. 

\begin{table}[t!]\scriptsize
\centering
\renewcommand\arraystretch{1.1}
\begin{tabular}{l|c|c}
\toprule[1pt]
\makecell{Method} & \makecell{PSNR$\uparrow$} & \makecell{SSIM$\uparrow$}   \\
\cline{1-3}
IBRnet\cite{wang2021ibrnet}      & 21.43    & 0.892        \\ 
Neuray\cite{liu2022neuray}      & 23.34    & 0.909          \\ 
NeuralHumanFVV\cite{9578599}      & 21.69    & 0.914         \\ 
NeuralHOIFVV     & 23.10     & 0.912      \\
\bottomrule[1pt]
\end{tabular}
\caption{\textbf{Neural rendering benchmark} in sparse-view setting.}
\label{tab: render}
\end{table}

 \subsection{Limitations} 


Although NeuralDome can provide accurate capturing results and layer-wise neural representations, we also want to highlight some potential limitations of this pipeline. First, NeuralDome only considers single-person interaction with objects and the single-person task is already quite challenging for current research. It is non-trivial to extend the current algorithm to multi-person settings, especially in a crowd scene. Secondly, the reconstruction of 3D holistic scene is not covered in our pipeline. We leave the joint modeling of humans, objects and scene in the future work. Moreover, our dataset was collected under fixed illumination conditions with few backgrounds variance, limiting its generalization ability for other environments. 


\section{Conclusion}

We have presented HODome, the first dataset to jointly
capture and render human-object interaction using dense RGB cameras and the Optitrack system. To process the HODome dataset, we have further developed NeuralDome, a neural pipeline that can accurately track humans and objects, conduct layer-wise geometry reconstruction, and enable novel-view synthesis for both the subjects and objects. We believe the dataset could boost the development of capturing and rendering human-object interactions and is very valuable to the community for a wide range of human-object capturing and rendering tasks.



\renewcommand\thesection{\Alph{section}}
\setcounter{section}{0}
\section{More Details of Data Capturing}

In this section, we discuss more details about our data capturing system and annotation process.

\subsection{Data Preprocess} 
Accurate human-object segmentation is required for joint optimization of capturing,  Instant-NSR~\cite{zhao2022human} and Instant-NGP~\cite{mueller2022instant}. For each recording frame, we segment the human-object foreground mask by running background matting \cite{sengupta2020background} and detect 2D joints (include hand) by using state-of-the-art 2D joint detectors, e.g. OpenPose \cite{cao2017realtime}.

\subsection{Object Tracking} 
For object tracking, each object is represented by a rigid mesh by pre-scanned tenplate. To capture object motions, we attach 10 mm hemispherical markers with strong glue directly to the object's surface and use at least 4 markers for each object. Note that we empirically distribute them on the object to ensure at least 3 of them are always observed.

\subsection{Time Synchronization} 
To avoid motion blur, we set the Optitrack system~\cite{Optitrack} to work at 120 FPS and set the RGB system to work at 60 FPS. Thus we need to resample the tracking data (120Hz) to the same frame rate as RGB (60Hz) to synchronize the RGB and marker system. Besides, we place an additional marker on the hand of the actor and block it as the flag of starting. Then we manually label the start frame in both recording timestamps. 

\subsection{Calibration}
To align RGB and Optitrack into the unified world coordinate system w.r.t. to the scene, we need to register the Optitrack maker set to the 3D scene. First human annotator annotates 3 correspondences between the marker recording data and RGB world location and then estimates the marker-to-RGB rigid transformation using ICP \cite{besl1992method,zhou2018open3d}. Besides, camera extrinsic parameters and rigid transformation are fixed during each recording.

\subsection{Data Capture Protocol}
We recruit 10 subjects (5 males and 5 females) that are between 18-40 years old and between 1.5-1.95m heights. Each subjects are recorded while interacting with 23 objects, according to their time availability. To ensure interact with objects as naturally as possible, each subjects are not instructed to do any actions. 



\section{Contribution to Community}
Consisting of various human-object interacting scenes with rich labels covering capturing and rendering labels, our NeuralDome pipeline and HODome dataset fill an important gap in the literature and support many research directions. We propose the following challenges with HODome dataset:

\myparagraph{Interaction Capturing.}
HODome dataset provided the largest accurate capturing label with paired natural RGB images for strong HOI supervision. Benefiting from our 76-view setting, our dataset is suitable for the monocular and multi-view settings. Moreover, our quantitative subset can be used for benchmarking thanks to our accurate ground truth and dense view validation.

\myparagraph{Geometry Reconstruction.}
Joint human-object geometry Reconstruction is a challenging task and the existing dataset can not be used for the benchmark. Besides, existing publicly available data do not support learning an accurate data-driven model of human-object geometry. Thanks to our dense capture setting, NeuralDome can provide high-fidelity human/object geometry to enable this task. 

\myparagraph{Object-Occluded Pose and Shape Estimation.}
Existing public datasets (e.g. Human36m~\cite{h36m_pami} and 3DPW~\cite{von2018recovering}) mainly focus on capturing accurate human labels but ignore the object conditions. By contrast, HODome dataset provided accurate capturing labels with SMPL parameters and 3D keypoints under challenge object-occluded case. 
 
\myparagraph{Neural Rendering in HOI scenarios.}
With the human-object interaction sequences captured by our dense cameras, there are lots of interesting and meaningful directions to explore. 
First, it's interesting to extend our layer-wise human-object representation to weaker settings that are closer to real life, e.g., sparse views, without accurate object poses.
Besides, generalizable neural rendering techniques should be further developed to support HOI scenarios where occlusions are inevitable.
Moreover, our HODome can naturally enable building photo-realistic neural avatars that support object-aware deformation in HOI scenarios. 






\section{More Details of Human-object Tracking}
In Section 4.1 of the paper, we have described the joint tracking between humans and objects. Here we elaborate more details and mathematical formulations. 


\subsection{Tracking Initialization} 

\myparagraph{Object Tracking.} 
We consider each object $V \in \mathbb{R}^{m}$ as a rigid body mesh model with $m$ vertices. And we only need to estimate the translation $T_t \in \mathbb{R}^{3}$ and rotation $R_t \in \mathbb{SO}(3)$ with respect to its pre-scanned template on each frame $t$. The 3D location of the object mesh on per-frame is represented as,
\begin{equation}
V_t(R_t,T_t) = R_t \mathcal{O}(c_t, p_t) + T_t,
\end{equation}
where $\mathcal{O}(c_j)$ represents the $p_j$ part of category $c_j$ mesh template. $T_t$ and  $R_t$ is rigid transformation estimated from a per-frame marker set using Rigid-ICP.

\subsection{Joint optimization for human-object tracking} 

We used SMPL-X \cite{pavlakos2019expressive} as the body
model, which provides a differentiable function $\mathcal{M}(\cdot)$ to control an artist-created mesh with $N=10475$ vertices and $K=54$ joints. We estimate body shape and pose over the whole sequence from multi-view RGB videos in a frame-wise manner. Recall that we imposed several regularization terms in Eq. (1) to ensure plausible interactions. Here we elaborate the mathematical formulation of each term.

Following the previous method~\cite{huang2022intercap,bhatnagar2022behave}, we compute the index of vertex where the body is in contact with objects, and enforce contact between SMPL-X and object explicitly as the following term:
\begin{equation}
\begin{aligned}
E_{\text{contact}}=  \| \mathbb{1}_t^{\text{contact}}( V_t(R_t,T_t) - \mathcal{M}(\beta_t,\theta_t,\psi_t, \gamma_t) ) \|_2^2 ,
\end{aligned}
\end{equation}
where $\mathbb{1}_t^{\text{contact}}$ denote as binary indicator matrix computed from the contact map. Note that marker-based tracking is quite accurate but remains some bias due to the error caused by camera alignment. Thus we impose a human-object silhouettes loss term using a differentiable render~\cite{kato2018neural} to refine the 3D object and SMPL-X to human-object foreground masks: 
\begin{equation}
\begin{aligned}
E_{\text{homask}}=  \| \sum_1^{76} ( I_j^{\text{homask}} - DR(V_t , \mathcal{M} )  \|_2^2 ,
\end{aligned}
\end{equation}
where $ DR $ denote as differentiable rendering and $I_j^{homask}$ denote as human-object mask computed from background matting. Finally, we also impose a marker corresponding team to prevent local minimal,

\begin{equation}
\begin{aligned}
E_{\text{marker}}=  \| \mathbb{1}^{\text{marker}} V_t(c_j, p_n) - s ) \|_2^2 ,
\end{aligned}
\end{equation}
where $ \mathbb{1}^{\text{marker}}$ denote as a binary indicator matrix that selects the vertices on the object mesh $V_t$ at its marker location and $s$ is the tracking data by the marker.



\section{More Details of NeuralHOIFVV}

\begin{figure}[t]
\includegraphics[width=1.0 \linewidth]{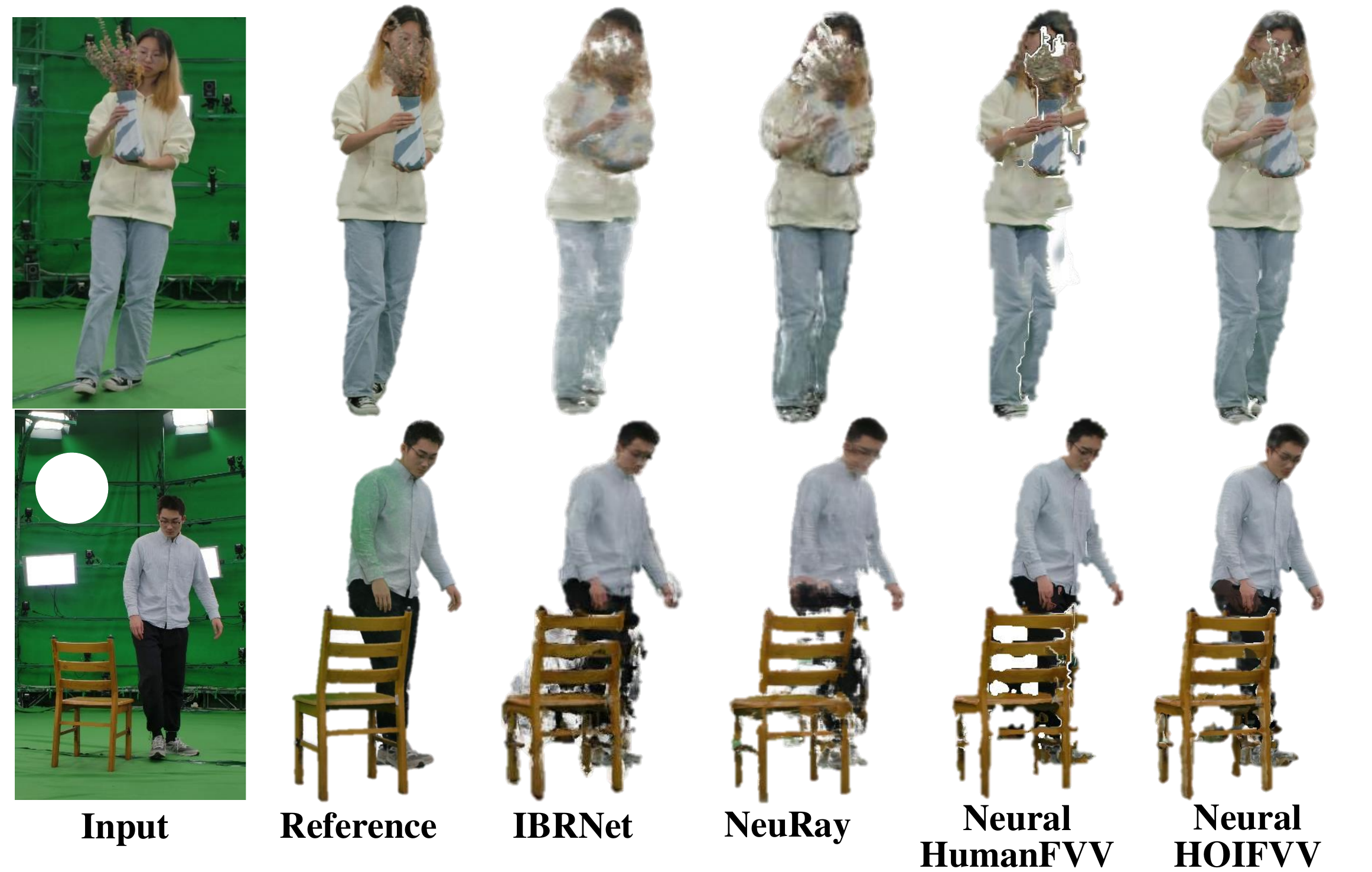}
\captionof{figure}{We provide more quality result on sparse-view rendering tasks and evaluate on IBRNet \cite{wang2021ibrnet}, NeuRay \cite{liu2022neuray}, NeuralHumanFVV \cite{9578599} and our baseline NeuralHOIFVV.}
\label{fig:neuralhoifvv}
\end{figure}

To validate that the HODome dataset is able to support novel view synthesis under sparse view settings, we propose a naive method called NeuralHOIFVV. We adapt the neural texture blending method introduced in \cite{zhao2022human}. Instead of using precise depth, we use the reprojection of 6-view PIFu \cite{saito2019pifu} trained on HODome to generate the coarse depth maps of the target view and input views. After getting the coarse depth map of the target view, we use it to warp the input images and input coarse depth maps into the target view. Then we use the same network as introduced in \cite{zhao2022human} to predict the two channels' feature maps $W = (W_1,W_2)$ representing the blending weights of warped images. Note that different from \cite{zhao2022human}, we do not have a coarse rendering image at the target view generated by the textured mesh. Our blending result is obtained only by using the blending map $W$ and the warped images. We provide also provide more quality result on sparse-view rendering tasks and evaluate on IBRNet \cite{wang2021ibrnet}, NeuRay \cite{liu2022neuray}, NeuralHumanFVV \cite{9578599} and our baseline NeuralHOIFVV as shown in Fig. \ref{fig:neuralhoifvv}.


\begin{figure}[t!]
\includegraphics[width=1.0 \linewidth]{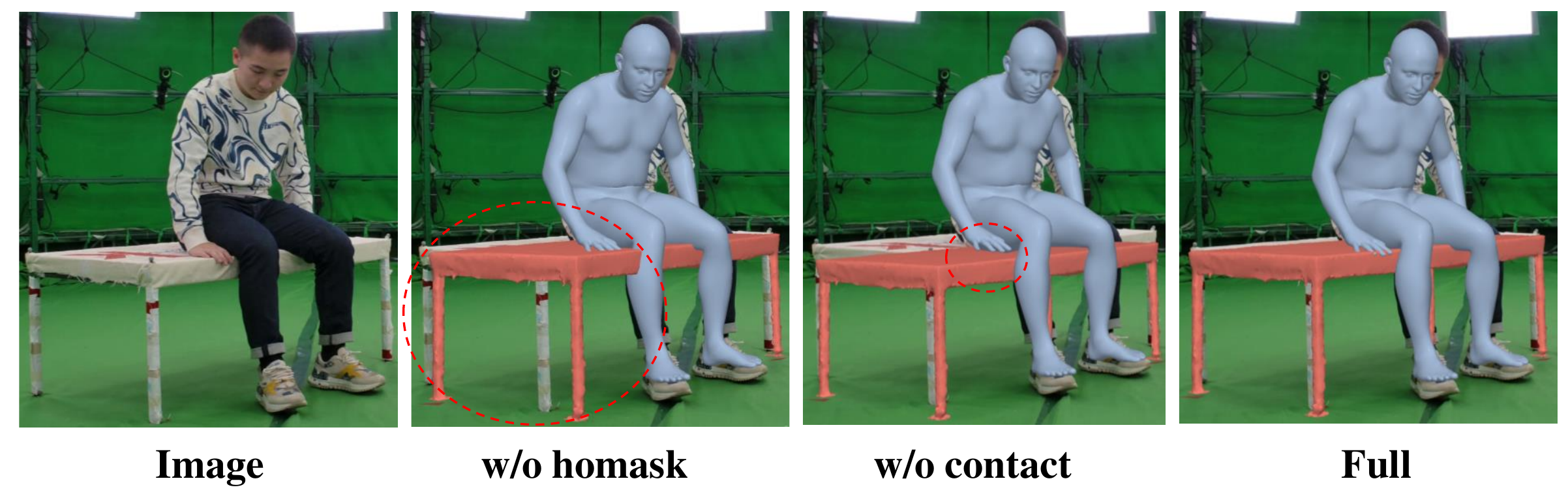}
\captionof{figure}{\textbf{Qualitative evaluation}}
\label{fig:joint}
\end{figure}

\begin{figure*}[t]
  \centering
  \includegraphics[width=1.0\linewidth]{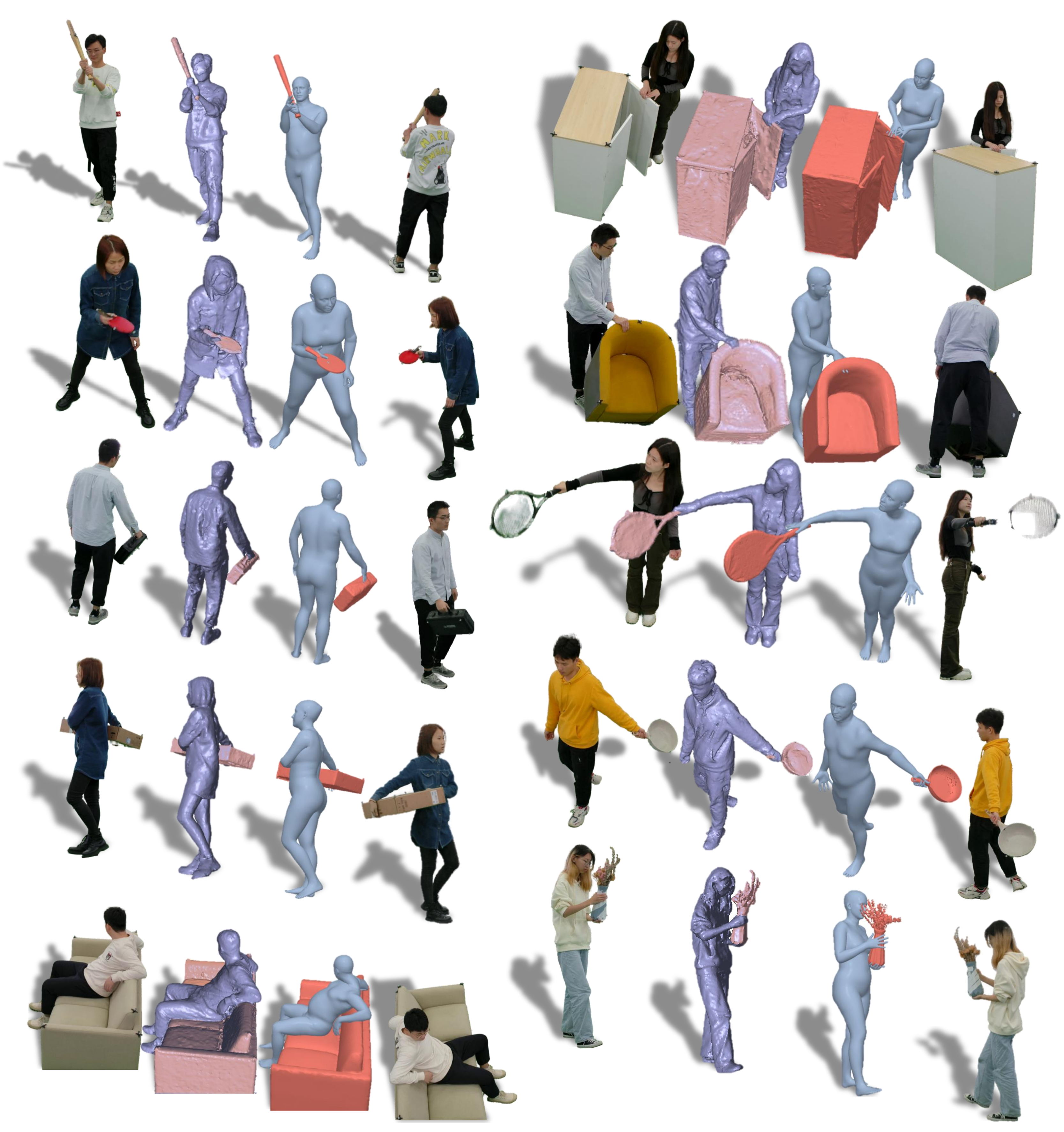}
 \caption{\textbf{More quality results.}}
  \label{fig:morequality}
\end{figure*}

\begin{figure*}[t]
  \centering
  \includegraphics[width=1.0\linewidth]{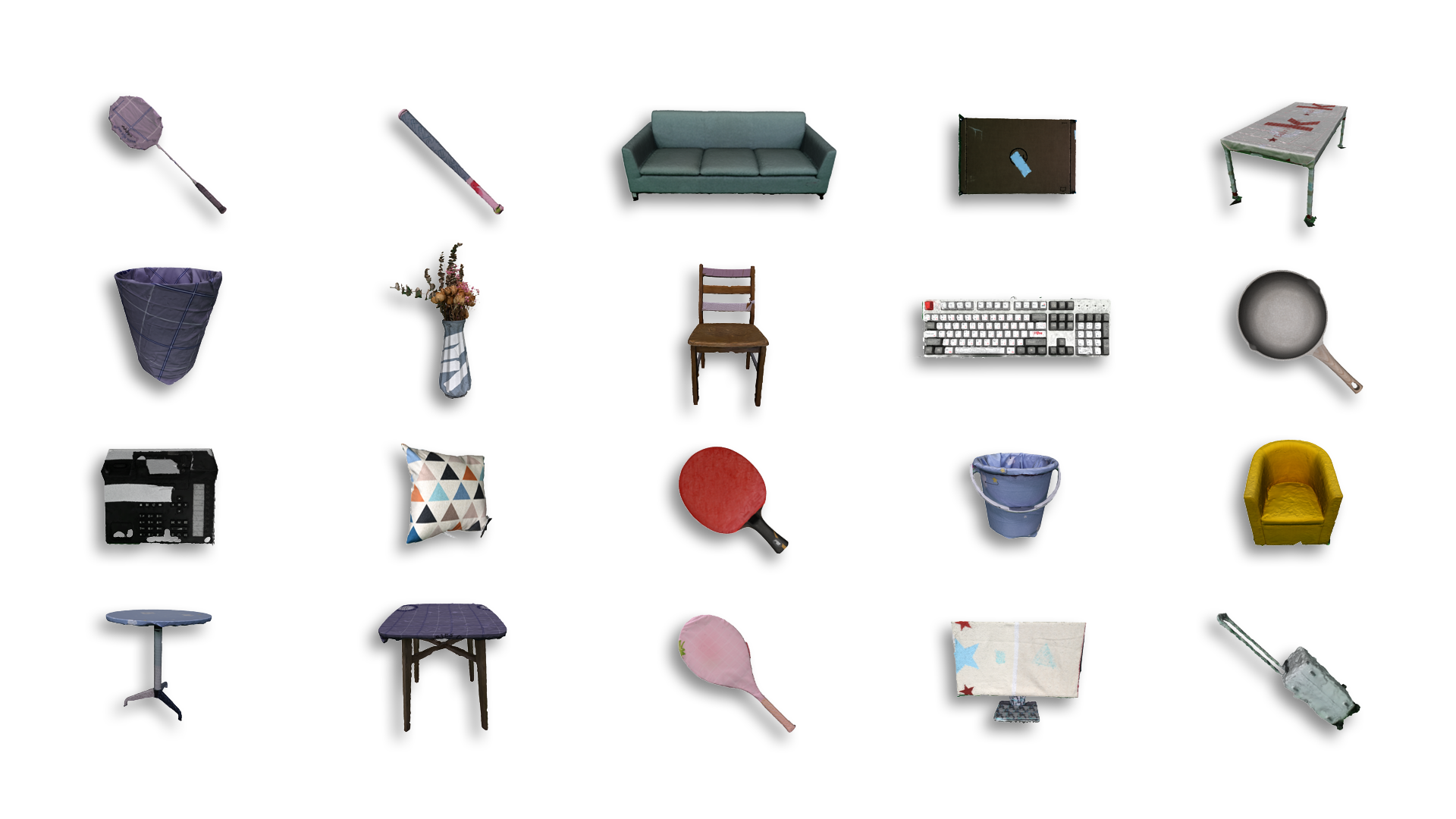}
 \caption{ The objects sampled from our HODome dataset}
  \label{fig:object}
\end{figure*}

\begin{figure*}[t]
  \centering
  \includegraphics[width=1.0\linewidth]{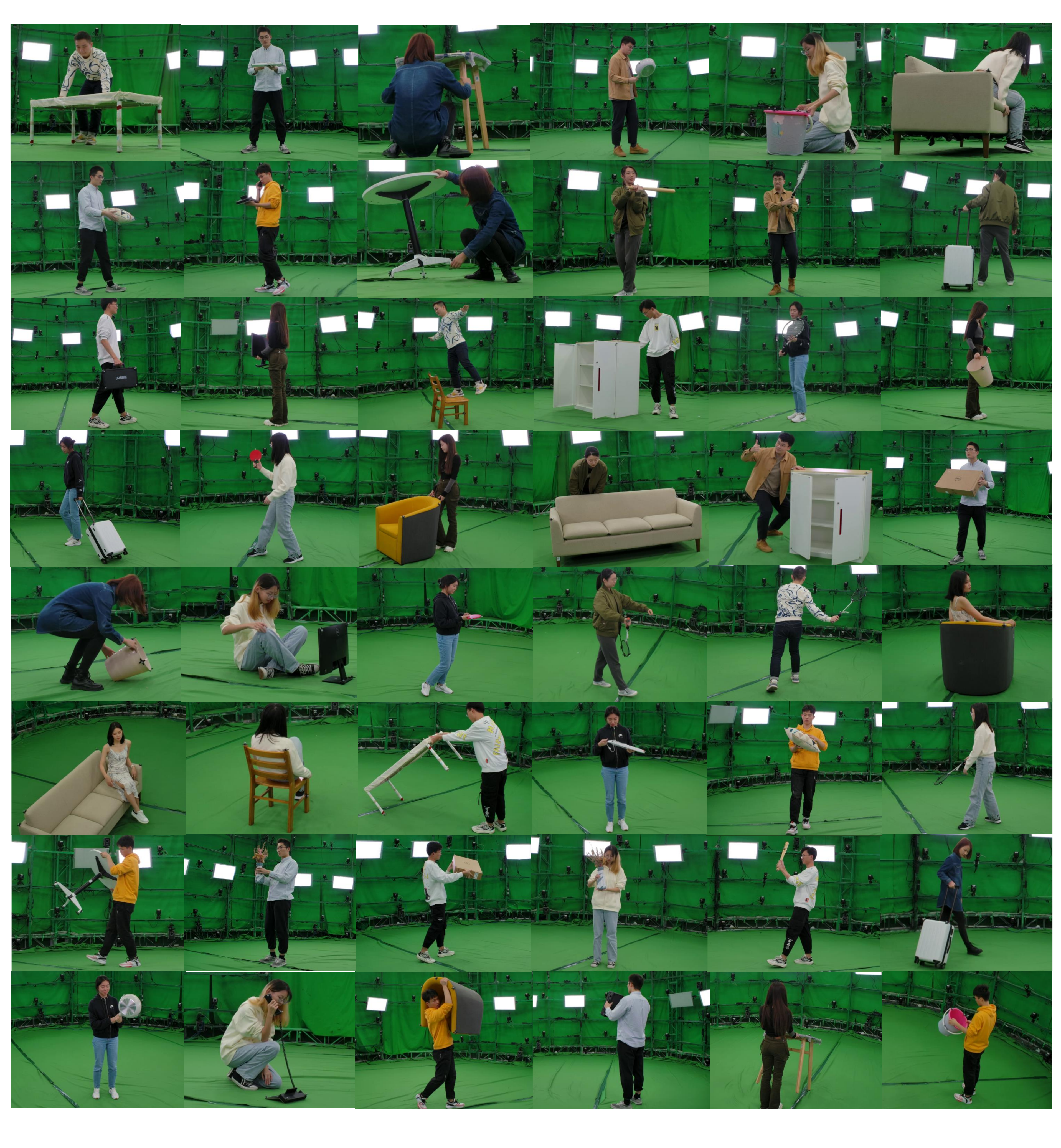}
 \caption{ Data examples were captured by our multi-view HODome with 76 synchronized high-resolution RGB cameras and Optitrack system. Our dataset includes a variety of human-object under various interactions}
  \label{fig:Dataexamples}
\end{figure*}

\section{More Experiment Results}

To better evaluate the components of joint optimization, we further do an additional quality analysis of different constraint terms. Note that we have no ground truth of the specific tracking, thus we conduct qualitative evaluations only. Fig.~\ref{fig:joint} shows the quality result by ablating different components. "w/o homask" and "w/o contact" respectively denotes the result obtained without using the human-object masks term $E_{\text{homask}}$, and without using the contact term $E_{\text{contact}}$. It demonstrates that the term of human-object masks $E_{\text{homask}}$ can effectively alleviate the bias caused by calibration and alignment and our contact map further ensures the realistic interaction between humans and objects. We also provide more quality results sampled from our HODome dataset as shown in Fig.~\ref{fig:morequality}. Fig.~\ref{fig:Dataexamples} provides the gallery of data examples captured by our multi-view HODome with 76 synchronized high-resolution RGB cameras and Optitrack system. Our dataset includes a variety of human-object under various interactions. Fig.~\ref{fig:object} shows the objects of data sampled from our HODome dataset.

{\small
\bibliographystyle{ieee_fullname}
\bibliography{egbib}
}

\end{document}